\documentclass{article}

\usepackage[accepted]{icml2019}

\usepackage[style=alphabetic,natbib=true,backend=biber,maxnames=10]{biblatex}
\bibliography{lml.bib}

\usepackage[utf8]{inputenc} %
\usepackage[T1]{fontenc}    %

\usepackage{color}
\definecolor{link_color}{RGB}{0,128,255}
\usepackage[%
colorlinks=true,allcolors=link_color,pageanchor=true,%
plainpages=false,pdfpagelabels,bookmarks,bookmarksnumbered,%
]{hyperref}

\usepackage{url}            %
\usepackage{booktabs}       %
\usepackage{amsfonts}       %
\usepackage{nicefrac}       %
\usepackage{microtype}      %
\usepackage{graphicx}
\graphicspath{{./images/}}

\usepackage{wrapfig}
\usepackage{subfigure}
\usepackage{booktabs} %
\usepackage{multirow}

\usepackage{algorithm}
\usepackage{algorithmicx}
\usepackage{algpseudocode}

\usepackage{todonotes}

\newcommand{\cblock}[3]{
  \hspace{-1.5mm}
  \begin{tikzpicture}
    [
    node/.style={square, minimum size=10mm, thick, line width=0pt},
    ]
    \node[fill={rgb,255:red,#1;green,#2;blue,#3}] () [] {};
  \end{tikzpicture}%
}

\usepackage{amsmath}
\usepackage{amsthm}
\usepackage{amssymb}
\usepackage{stmaryrd}

\newcommand{\nwc}{\newcommand}
\nwc{\as}{\textrm{a.s.}}
\nwc{\defas}{:=}

\nwc{\dist}{\ \sim\ }
\nwc{\distiid}{\stackrel{\mathrm{iid}}{\sim}}

\DeclareMathOperator*{\argmin}{argmin}

\DeclareMathOperator*{\st}{s.t.}

\newcommand{\U}{\mathcal{U}}

\newcommand{\x}{\bm{x}}
\newcommand{\D}{\mathcal{D}}
\newcommand{\X}{\mathcal{X}}
\newcommand{\Y}{\mathcal{Y}}
\newcommand{\s}{\bm{s}}

\newcommand{\E}{\mathop{\mathbb{E}}}
\newcommand{\f}{\bm{f}}
\newcommand{\F}{\bm{F}}

\newcommand{\K}{\bm{K}}

\newcommand{\JJ}{\mathcal{J}}

\newcommand{\LML}{\ensuremath{\mathcal{L}}}

\newcommand{\RR}{\mathbb{R}}

\usepackage{accents}

\newtheorem{theorem}{Theorem}

\newtheorem{proposition}{Proposition}
\newtheorem{corollary}{Corollary}

\usepackage{lipsum}

\usepackage[capitalise,nameinlink,noabbrev]{cleveref}

\newcounter{module}
\makeatletter
\newenvironment{module}[1][htb]{%
  \let\c@algorithm\c@module
    \renewcommand{\ALG@name}{Module}%
   \begin{algorithm}[#1]%
  }{\end{algorithm}}
\makeatother
\crefname{module}{Module}{Modules}

\usepackage{xspace}
\newcommand{\eg}{{\it e.g.}\xspace}

\icmltitlerunning{The Limited Multi-Label Projection Layer}

\begin{document}

\twocolumn[
\icmltitle{The Limited Multi-Label Projection Layer}

\icmlsetsymbol{intern}{*}

\begin{icmlauthorlist}
  \icmlauthor{Brandon Amos}{cmu,intern}
  \icmlauthor{Vladlen Koltun}{intel}
  \icmlauthor{J.~Zico Kolter}{cmu,bosch}
\end{icmlauthorlist}

\icmlaffiliation{cmu}{Carnegie Mellon University}
\icmlaffiliation{intel}{Intel Labs}
\icmlaffiliation{bosch}{Bosch Center for AI}

\vskip 0.3in
]

\printAffiliationsAndNotice{\textsuperscript{*}Work done while BA was an intern at Intel Labs.\hspace{10mm}}

\begin{abstract}
We propose the Limited Multi-Label (LML) projection
layer as a new primitive operation for end-to-end learning systems.
The LML layer provides a probabilistic way of modeling
multi-label predictions limited to having exactly $k$ labels.
We derive efficient forward and backward passes for this layer
and show how the layer can be used to optimize the top-$k$
recall for multi-label tasks with incomplete label information.
We evaluate LML layers on top-$k$ CIFAR-100 classification and
scene graph generation. We show that LML layers add a negligible
amount of computational overhead, strictly improve the
model's representational capacity, and improve accuracy.
We also revisit the truncated top-$k$ entropy method as a
competitive baseline for top-$k$ classification.
\end{abstract}

\section{Introduction}

Multi-label prediction tasks show up frequently
in computer vision and language processing.
Multi-label predictions can arise from a task being truly
multi-label, as in language and graph generation tasks,
or by turning a single-label prediction task into a multi-label
prediction task that predicts a set of top-$k$ labels,
for example.
In high-dimensional cases, such as scene graph generation,
annotating multi-label data is difficult and often results
in datasets that have an incomplete labeling.
In these cases, models are typically limited to predicting
$k$ labels and are evaluated on the recall,
the proportion of known labels that are present in
the model's predicted set.
As we will show later, the standard approaches of using
a softmax or sigmoid functions
are not ideal here as they have no way of allowing the model
to capture labels that are unobserved.

In this report, we present the LML layer as a new way of
modeling in multi-label settings where the model needs to make a
prediction of exactly $k$ labels.
We derive how to efficiently implement and differentiate
through LML layers in \cref{sec:lml:lml}.
The LML layer has a probabilistic interpretation and can be
trained with a standard maximum-likelihood approach
that we show in \cref{sec:lml:topk}, where we also
highlight applications to top-$k$ image classification
and scene graph generation.
We show experiments in \cref{sec:lml:ex} on
\mbox{CIFAR-100} classification and
scene graph generation.

\section{Background and Related Work}
\begin{figure}[t]
  \centering
  \includegraphics[width=.22\textwidth]{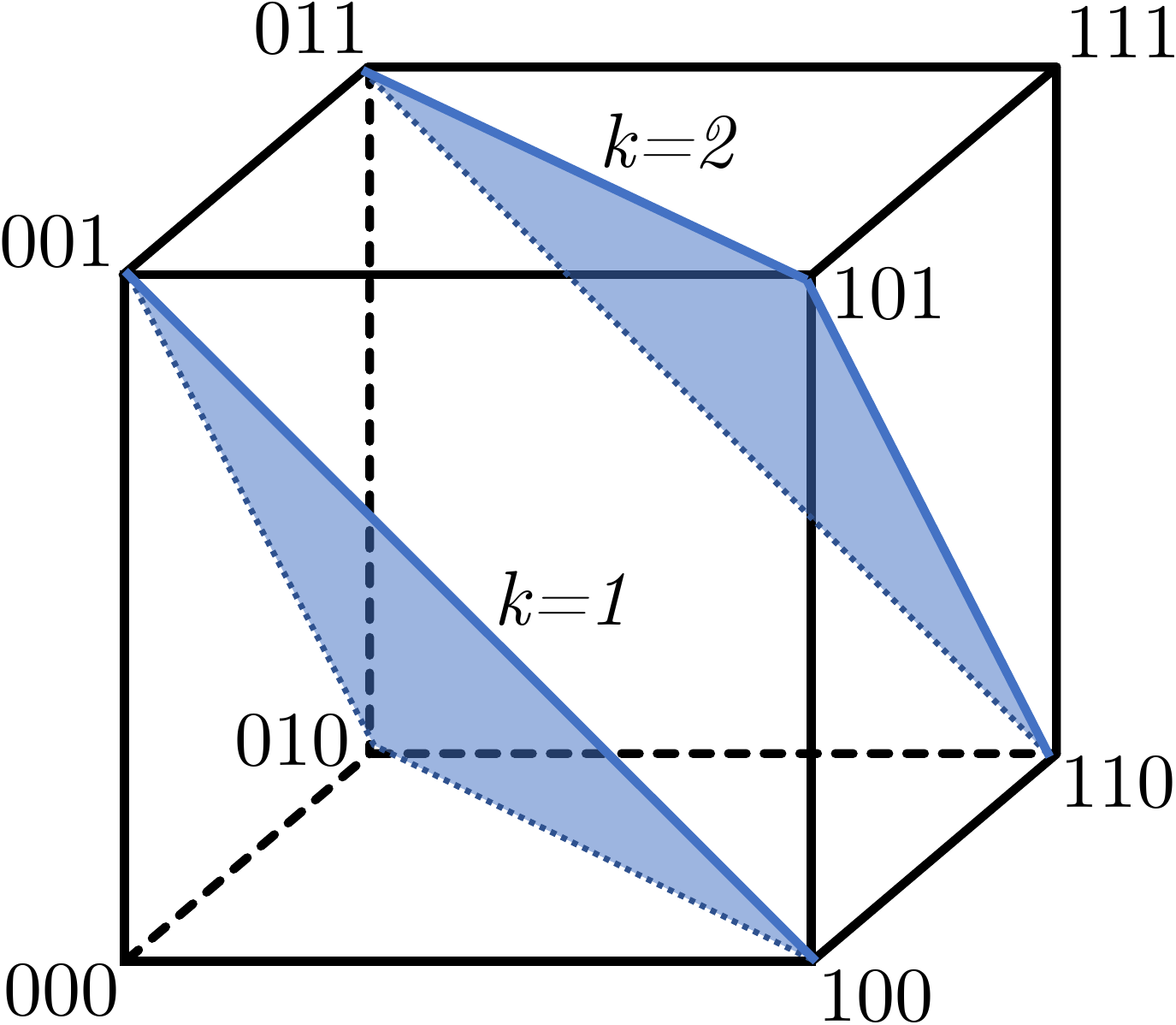}
  \includegraphics[width=.24\textwidth]{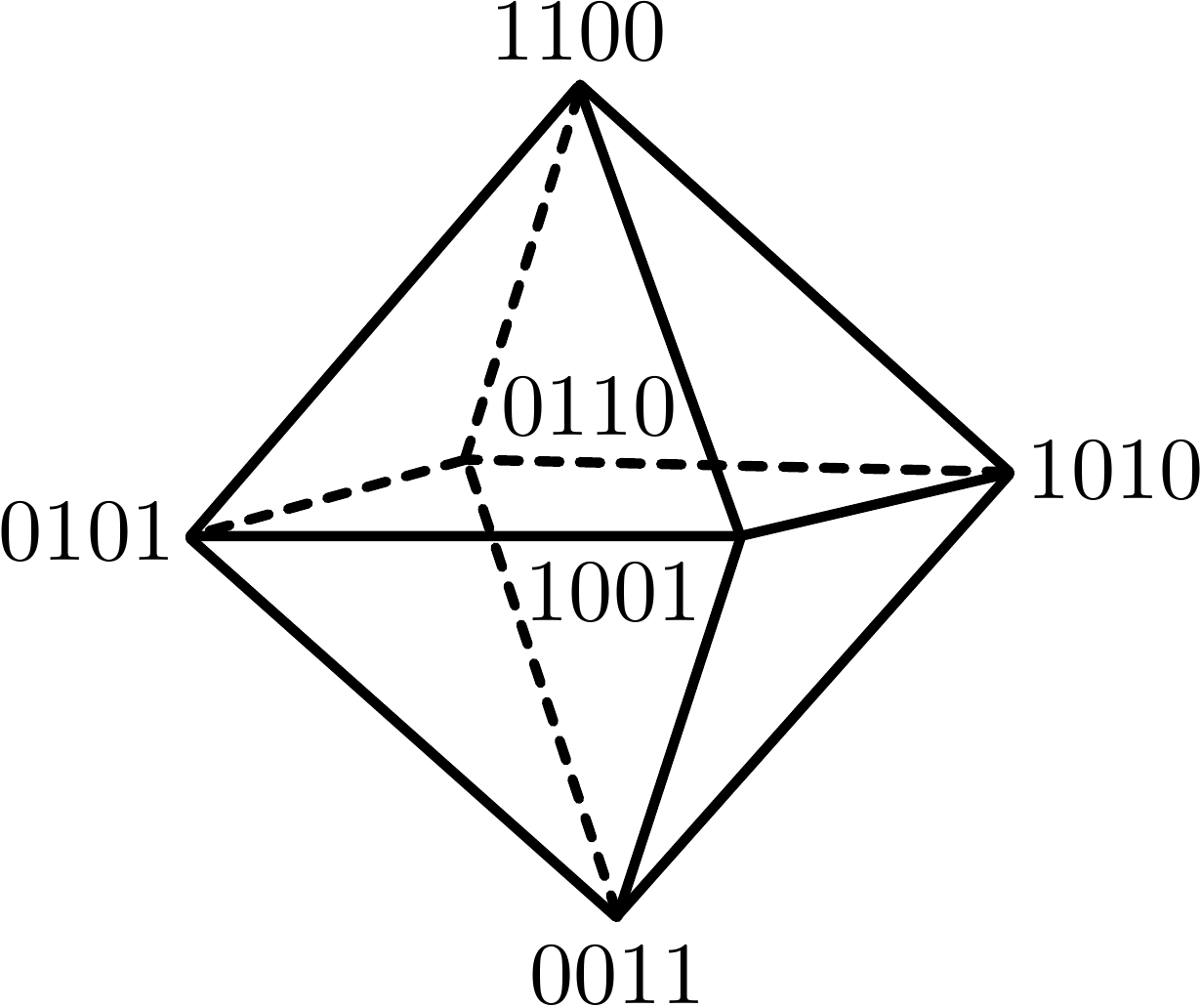}
  \caption{
    The LML polytope $\LML_{n,k}$ is the set of points in the unit
    $n$-hypercube with coordinates that sum to $k$.
    $\LML_{n,1}$ is the \mbox{$(n-1)$-simplex}.
    The $\LML_{3,1}$ and $\LML_{3,2}$ polytopes (triangles) are on the left
    in blue. The $\LML_{4,2}$ polytope (an octahedron) is on the right.
  }
  \label{fig:lml-example}
\end{figure}

\subsection{Differentiable Projections in Machine Learning}
\label{sec:bg:projs}

Differentiable projections onto polytopes are ubiquitous in
machine learning.
Many standard operations can be interpreted as projections
onto polytopes, such as the sigmoid, softmax, and ReLU, as
described, \eg, in \cite[Section 2.4]{amos2019differentiable}.
Similar projections are also done onto more complex polytopes
such as the marginal polytope for structured inference
\citep{niculae2018sparsemap}
or the Birkhoff polytope for permutations
\citep{adams2011ranking,santa2018visual,mena2018learning}.
Very closely related to our work is the \emph{constrained softmax}
proposed in \cite{martins2017learning} that is studied in
the context of sequence tagging and the \verb!amx! layer
studied in \cite{globerson2016collective} for collective
entity resolution.
As a special case, the constrained softmax can be used to do
an entropy-based projection onto the
\emph{capped simplex} \citep{warmuth2008randomized}.
Under a change of variables described, \eg, in
\citep[Appendix C.2]{blondel2019structured},
the constrained softmax projects onto the same set
that we consider and also shows how to differentiate through
this projection for learning.
A special case of the \verb!amx! layer when $\beta=1$ takes the min
of this projection instead of the argmin.
\cite{malaviya2018sparse} proposes the \emph{constrained sparsemax}
that uses a quadratic penalty instead of an entropy penalty.

The constrained softmax can likely
be used in every scenario we consider the LML projection
for in this paper, as the forward and backward passes provide
similar signals for learning.
The only difference between our LML projection and the special
case of the constrained softmax is the entropy penalty used
over the LML polytope --- the constrained softmax uses a
unidirectional entropy term and the LML projection uses a
binary entropy term.
We visualize these surfaces in \cref{app:entropies} to
motivate our choice of the binary entropy function.

\subsection{Cardinality Potentials and Modeling}
Cardinality potentials and modeling are a closely related line
of work typically found in the structured prediction
and constraint programming literature.
\citet{regin1996generalized} shows how to add constraints to
models for worker scheduling.
\citet{tarlow2012fast} propose a way of performing structured
prediction with cardinality potentials, and \citet{brukhim2018predict}
propose a soft projection operation that integrate
cardinality modeling into deep structured prediction
architectures like SPENs \citep{belanger2015structured}.
In contrast to these methods, our projection and
constraint is exact and can be integrated in the standard
forward pass of a deep model outside of structured prediction.
None of our experiments use structured prediction techniques
and we instead do standard supervised learning of vanilla
feedforward models that use our LML layer.
In contrast to \citet{brukhim2018predict}, we show that
the backward pass of our soft projection can be exactly
computed instead of unrolled as part of a structured
prediction procedure.

\subsection{Top-$k$ and Ranking-Based Loss Functions}
\label{sec:lml:rw:topk}
There has been a significant amount of work on creating
specialized loss functions for optimizing the model's
top-$k$ prediction error
\citep{gupta2014training,li2014top,liu2015transductive,
  lapin2015top,liu2015transductive,lapin2016loss,berrada2018smooth}
and ranking error
\citep{agarwal2011infinite,rudin2009p,boyd2012accuracy,rakotomamonjy2012sparse}.

Most relevant to our contributions are the smooth
top-$k$ loss functions discussed in
\citet{lapin2016loss} and the Smooth SVM \citep{berrada2018smooth}.
Among other loss functions, \citet{lapin2016loss} propose the
truncated top-$k$ entropy loss,
which we review in \cref{sec:lml:entr-derivation} and
extend to cases when multiple ground-truth labels are
present in \cref{sec:lml:entr-ml-derivation}.

In contrast to all of these methods, our approach does
not hand-craft a loss function and instead puts
the top-$k$ knowledge into the modeling part of the
pipeline, which is then optimized as a likelihood
maximization problem.
We show in \cref{sec:lml:cifar} that LML layers are competitive
in the top-$k$ prediction task from \citet{berrada2018smooth}.

\newpage
\subsection{Scene Graph Generation}
\label{sec:lml:rw:sg}

Scene graph generation is the task of generating a set
of objects and relationships between them from an
input image and has been extensively studied recently
\citep{johnson2015image,yang2017support,plummer2017phrase,
  liang2017deep,raposo2017discovering,newell2017pixels,
  xu2017scene,li2018factorizable,
  herzig2018mapping,zellers2018neural,woo2018linknet}.
Most relevant to our work are the methods that score
all of the possible relationships between objects and
select the top-scoring relationships
\citep{xu2017scene,li2018factorizable,herzig2018mapping,woo2018linknet}.
These methods include the near-state-of-the-art
Neural Motifs model \citep{zellers2018neural}
that generates a scene graph by creating object-
and edge-level contexts.

We propose a way of improving the relationship prediction
portion of methods that fully enumerate all of the
possible relationships, and we empirically
demonstrate that this improves the
representational capacity of Neural Motifs.

\section{The Limited Multi-Label Projection Layer}
\label{sec:lml:lml}

We propose the \emph{Limited Multi-Label projection layer} as
a way of projecting onto the set of points in the unit
$n$-hypercube with coordinates that sum to exactly $k$.
This space can be represented as a polytope, which we define
as the \emph{(n,k)-Limited Multi-Label polytope}
$$\LML_{n,k} = \{p\in\RR^n \mid 0\leq p\leq 1 \;\;
  {\rm and} \;\; 1^\top p = k\}.$$
When $k=1$, the LML polytope is the $(n-1)$-simplex.
Notationally, if $n$ is implied by the context we will leave
it out and write $\LML_k$.
\cref{fig:lml-example} shows three low-dimensional examples
of this polytope.
The LML polytope is a scaled version of the capped simplex
studied on the context of online PCA in \cite{warmuth2008randomized}
and is an instance of the \emph{knapsack polytope} defined,
\eg, in \cite{blondel2019structured}.
A useful variant of this polytope that we do not consider
is the \emph{budget polytope} \citep{almeida2013fast},
which can capture the constraint $1^\top p\leq k$ instead
of $1^\top p= k$.

We consider projections onto the interior of the LML polytope
of the form
\begin{equation}
  \Pi_{\LML_k}(x) = \argmin_{0<y<1} \;\; -x^\top y - H_b(y) \;\; \st\;\; 1^\top y = k
  \label{eq:lml-proj}
\end{equation}
where $H_b(y) = - \left(\sum_i y_i\log y_i + (1-y_i)\log (1-y_i)\right)$
is the binary entropy function.
The entropy-based regularizer in the objective helps prevent
sparsity in the gradients of this projection, which
is important for learning and the same reason it is useful
in the softmax.
We note that other projections could be done by changing the
regularizer or by scaling the entropy term with
a temperature parameter, as done in the
\emph{constrained softmax} \citep{martins2017learning}
and
\emph{constrained sparsemax} \citep{malaviya2018sparse}.

\newpage
The following is one useful property of the LML projection
when $x$ is the output of a function such as a neural network.

\begin{proposition}\label{prop:order}
  $\Pi_{\LML_k}(x)$ preserves the (magnitude-based) order of
  the coordinates of $x$.
\end{proposition}
The intuition is that $\Pi_{\LML_k}(x)$ can be decomposed
to applying a monotonic transformation
to each element of $x$, which we show in \cref{eq:lml-fwd}.
Thus, this preserves the (magnitude-based) ordering of $x$.

The LML projection layer does not have an explicit closed-form
solution like the layers discussed in \cref{sec:bg:projs},
despite the similarity to the softmax layer.
We show how to efficiently solve the optimization problem
for the forward pass in \cref{sec:lml:lml:efficient}
and how to backpropagate through the LML projection in
\cref{sec:lml:lml:backprop} by implicitly differentiating
the KKT conditions.
\cref{mod:lml} summarizes the
implementation of the layer.

\begin{module}[t]
\caption{The Limited Multi-Label Projection Layer}
\label[module]{mod:lml}
\textbf{Input:} $x\in\RR^n$, $k\in\mathbb{N}$ \\
\textbf{Forward Pass} \hfill \emph{(Described in \cref{sec:lml:lml:efficient})}
\begin{algorithmic}[1]
  \State Compute $\nu^\star$ with \cref{alg:dual}
  \State \Return $y^\star = \sigma(x+\nu^\star) \in \LML_k$
\end{algorithmic}~\\
\textbf{Backward Pass} \hfill \emph{(Described in \cref{sec:lml:lml:backprop})}
\begin{algorithmic}[1]
  \State $h=(y^\star)^{-1}+(1-y^\star)^{-1}$
  \State $d_\nu = (1^\top h^{-1})^{-1} h^{-\top}\left(\nabla_{y^\star} \ell\right)$
  \State $d_y = h^{-1}\circ(d_\nu-\nabla_{y^\star}\ell)$
  \State \Return $\nabla_x \ell = -d_y$
\end{algorithmic}
\end{module}

\subsection{Efficiently computing the LML projection}
\label{sec:lml:lml:efficient}
The LML projection in \cref{eq:lml-proj} is a convex and
constrained optimization problem. In this section we
propose an efficient way of solving it that is GPU-amenable.

\begin{figure*}[ht!]
  \centering
  \hfill
  \includegraphics[width=.37\textwidth]{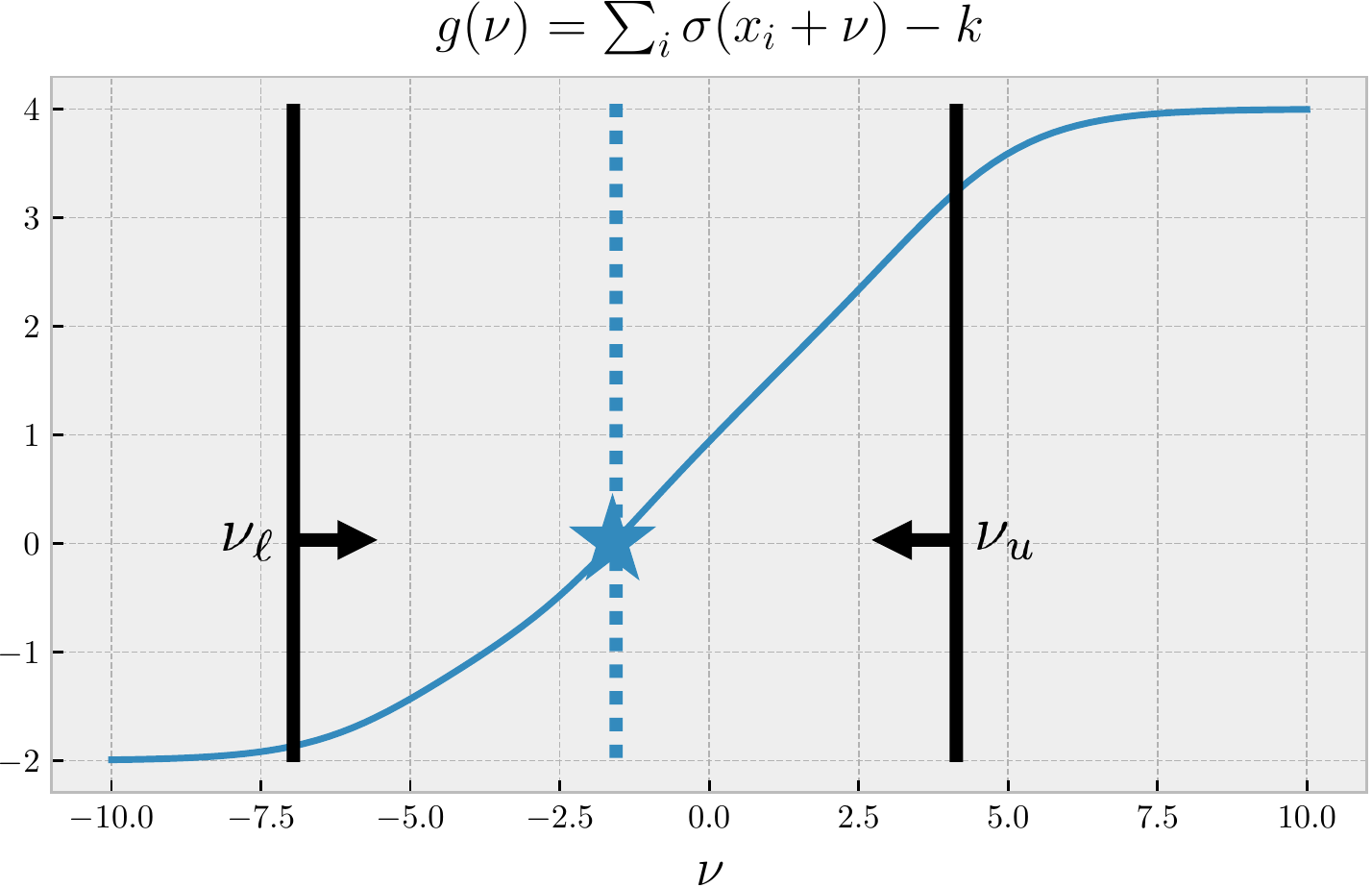}
  \hfill
  \includegraphics[width=.49\textwidth]{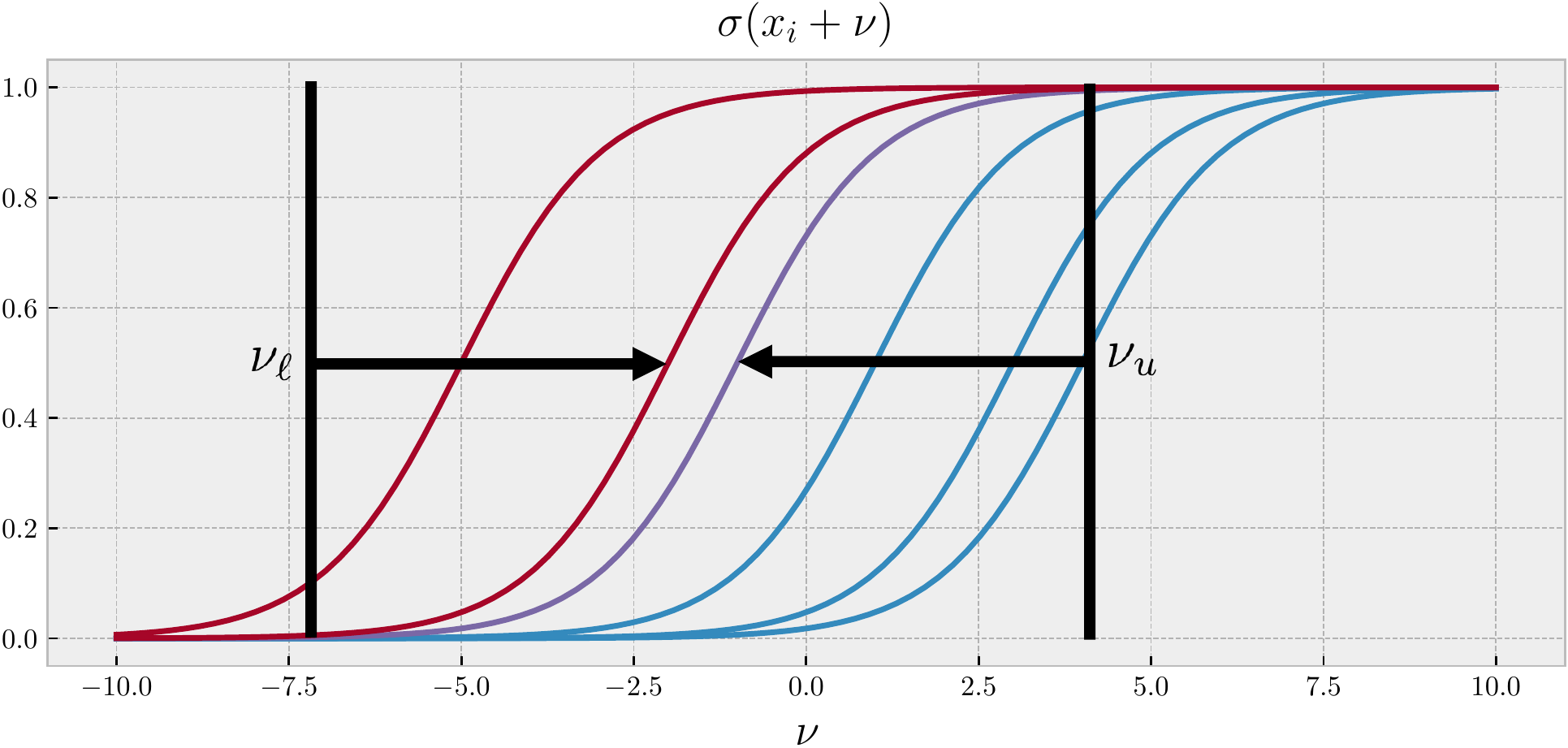}
  \hspace{-23mm}
  \begin{minipage}{2cm}%
  \vspace{-25mm} \footnotesize
  \hspace{0.1pt}\cblock{187}{64}{60} $\pi(x)_{1:k}$ \\
  \cblock{159}{92}{149} $\pi(x)_{k+1}$ \\
  \cblock{83}{123}{164} $\pi(x)_{k+2:n}$
  \end{minipage}
  \hfill

  \caption{
    Example of finding the optimal dual variable $\nu$
    with $x\in\RR^6$ and $k=2$ by solving the
    root-finding problem $g(\nu)=0$
    in \cref{eq:lml-root},
    which is shown on the left.
    The right shows the decomposition of the individual
    logistic functions that contribute to $g(\nu)$.
    We show the initial lower and upper bounds
    described in \cref{sec:lml:lml:dual}.
  }
  \label{fig:lml-root-ex}
\end{figure*}

Introducing a dual variable $\nu\in\RR$ for the constraint
\mbox{$k-1^\top y = 0$},
the Lagrangian of \cref{eq:lml-proj} is
$$L(y, \nu) = -x^\top y - H_b(y) + \nu(k-1^\top y),$$
where we unconventionally negate the equality constraint
to make analyzing $g(\nu)$ easier.
Differentiating this gives
\begin{equation}
\nabla_y L(y,\nu) = -x+\log \frac{y}{1-y} - \nu
\end{equation}
and first-order optimality
$\nabla_y L(y^\star,\nu^\star) = 0$
gives
\begin{equation}
  \label{eq:lml-fwd}
  y^\star=\sigma(x+\nu^\star),
\end{equation}
where $\sigma$ is the logistic function.
To find the optimal dual $\nu^\star$, we can
substitute \cref{eq:lml-fwd} into the constraint
\begin{equation}
  \label{eq:lml-root}
  g(\nu) \triangleq 1^\top \sigma(x+\nu) - k = 0.
\end{equation}
Thus the LML projection can be computed by
solving $g(\nu)=0$ for the optimal dual variable
and then using \cref{eq:lml-fwd} for the projection.

\subsubsection{Solving $g(\nu)=0$}
\label{sec:lml:lml:dual}

\begin{algorithm}[t]
  \caption{Bracketing method to find $g(\nu)=0$}
  \label{alg:dual}
  \textbf{Input:} $x\in\RR^n$ \\
  \textbf{Parameters:} $d$: the number of per-iteration samples \\
  \hspace*{18.5mm}$\Delta$: the saturation offset
  \begin{algorithmic}[1]
    \State Initialize $\nu_\ell=-\pi(x)_k-\Delta$ and $\nu_u=-\pi(x)_{k+1}+\Delta$
    \While {$|\nu_\ell - \nu_u| > \epsilon$}
    \State Sample $\nu_{1:d}$ linearly from the interval $[\nu_\ell, \nu_u]$
    \State $g_{1:d}=(g(\nu_i))_{i=1}^d$ \Comment \emph{Ideally parallelized}
    \State \(\triangleright\) \emph{Return the corresponding
        $\nu_i$ early if any $g_i=0$}
    \State $i_\ell = \max\{ i \mid g_i < 0\}$ and $i_u = i_\ell+1$
    \State $\nu_\ell = \nu_{i_\ell}$ and $\nu_u = \nu_{i_u}$
    \EndWhile
    \State \Return $(\nu_\ell+\nu_u)/2$
  \end{algorithmic}
\end{algorithm}

$g(\nu)=0$ is a scalar-valued root-finding problem
of a differentiable, continuous, non-convex function
that is monotonically increasing.
Despite the differentiability,
we advocate for solving $g(\nu)=0$ with a
\emph{bracketing method} that maintains an interval
of lower and upper bounds around the solution $\nu^\star$ and
is amenable to parallelization,
instead of a Newton method that would use the derivative
information but is not as amenable to parallelization.
Our method generalizes the bisection bracketing method
by sampling $g(\nu)$ for $d$ values of $\nu$ per iteration
instead of a single point.
On the GPU, we sample $d=100$ points in parallel
for each iteration, which usually reaches machine epsilon
in less than 10 iterations, and on the CPU we
sample $d=10$ points.
We present our bracketing method in \cref{alg:dual}
and show an example of $g(\nu)$ and the
component functions in \cref{fig:lml-root-ex}.

The initial lower bound $\nu_\ell$ and upper bound $\nu_u$ on the
root can be obtained by observing that $g(\nu)$ takes a sum of
logistic functions that are offset by the entries of $x\in\RR^n$
as $\sigma(x_j+\nu)$.
With high probability, we can use the saturated
areas of the logistic functions to construct the initial bounds.

Let $\pi(x)$ sort $x\in\RR^n$ in descending order so that
$$\pi(x)_1 \geq \pi(x)_2 \geq \ldots \geq \pi(x)_n$$
and $\Delta$ be a sufficiently large offset that causes
the sigmoid units to saturate.
We use $\Delta=7$ in all of our experiments.

Use $\nu_\ell = -\pi(x)_k-\Delta$ for the \textbf{initial lower bound.}
This makes $\sigma(x_j+ \nu_\ell)\approx 0$ for $x_j\in \pi(x)_{k,\ldots,n}$
and $0<\sigma(x_j+ \nu_\ell)<1$ for $x_j\in \pi(x)_{1,\ldots,k-1}$,
and thus $g(\nu_\ell) \leq -1 \leq 0$.

Use $\nu_u = -\pi(x)_{k+1}+\Delta$ for the \textbf{initial upper bound.}
This makes $\sigma(x_j + \nu_u)\approx 1$ for every
$x_j \in \pi(x)_{1,\ldots,k+1}$ and thus
$g(\nu_u) \geq 1 \geq 0$.

\subsection{Backpropagating through the LML layer}
\label{sec:lml:lml:backprop}
Let $y^\star = \Pi_{\LML_k}(x)$ be outputs of the LML layer
from \cref{eq:lml-proj}.
Integrating this layer into a gradient-based
end-to-end learning system requires that we
compute the derivative
$$
\frac{\partial\ell}{\partial x} =
\frac{\partial\ell}{\partial y^\star}
\frac{\partial y^\star}{\partial x},
$$
where $\ell$ is a loss function.
The LML projection $\Pi_{\LML_k}(x)$ does not have an explicit
closed-form solution and we therefore cannot use an
autodiff framework to compute the gradient
$\partial y^\star/\partial x$.
We note that even though the solution can be represented as
$y^\star = \sigma(x+\nu^\star)$,
differentiating this form is still difficult because
$\nu^\star$ is also a function of $x$.
We instead implicitly differentiate the KKT conditions
of \cref{eq:lml-proj}.
Using the approach described, e.g., in OptNet \citep{amos2017optnet},
we can solve the linear system
\begin{equation}
\label{eq:lml-kkt}
\begin{bmatrix}
H & -1 \\
-1^\top & 0 \\
\end{bmatrix}
\begin{bmatrix}
d_y \\ d_\nu
\end{bmatrix}
=
-
\begin{bmatrix}
\nabla_{y^\star}\ell \\ 0
\end{bmatrix}
\end{equation}
where $H=\nabla^2_y L(y,\nu)$ is defined by $H={\rm diag}(h)$ and
\begin{equation}
  h=\frac{1}{y^\star}+\frac{1}{1-y^\star}.
\end{equation}
The system in \cref{eq:lml-kkt} can be solved
analytically with
\begin{equation}
  d_\nu = \frac{1}{1^\top h^{-1}} h^{-\top}\left(\nabla_{y^\star} \ell\right)
  \quad
  {\rm and}
  \quad
  d_y = h^{-1}\circ(d_\nu -\nabla_{y^\star}\ell)
\end{equation}
where $\circ$ is the elementwise product
and $h^{-1}$ is the elementwise inverse.
Finally, we have that
$\nabla_x \ell = -d_y$.

\newpage
\section{Maximizing Top-$k$ Recall via
  Maximum Likelihood with The LML layer}
\label{sec:lml:topk}

In this section, we highlight one application of the
LML layer for maximizing the top-$k$ recall.
We consider a multi-label classification setting
where the data has an incomplete (strict) subset of
the true labels and we want to model the task by
predicting a set of exactly $k$ labels.
This setting comes up in practice for
predicting the top-$k$ labels in image classification
and in predicting a set of $k$ relationships in
a graph for scene graph generation,
which we discuss in
\cref{sec:lml:topk:image,sec:lml:topk:sg},
respectively.

Formally, we have samples $(x_i, Y_i)\sim \D$ from some data
generating process $\D$ with features $x_i\in\X_i $ and labels
$Y_i\subseteq Y^\star_i\subseteq\Y\triangleq\{1,\ldots,n\}$,
where $Y^\star_i$ are the
ground-truth labels and $Y_i$ are the \emph{observed} labels.
There is typically some $k\ll n$ such
that $|Y_i^\star|\leq k$ for all $i$.
We will model this by predicting exactly $k$ labels
$\hat Y_i \subseteq \{1, \ldots n\}$ where $|\hat Y_i|=k$.

The model's predictions should have high recall
on the observed data, which for a single sample
is defined by
\begin{equation*}
{\rm recall}(Y, \hat Y) =
  \frac{1}{|Y|} \sum_{j\in Y} \llbracket y_j\not\in \hat Y \rrbracket,
\end{equation*}
where the Iverson bracket $\llbracket P \rrbracket$
is 1 if $P$ is true and 0 otherwise.
We note that the 0-1 error, defined as
\begin{equation*}
  {\rm error}(Y, \hat Y) = \llbracket Y \neq \hat Y \rrbracket,
\end{equation*}
or smooth variants thereof, are not a reasonable proxy for
the recall as it incorrectly penalizes the model when
it makes a correct prediction $\hat Y$ that is in the ground
truth labels $Y^\star$ but not in the observation $Y$.

We will next use a probabilistic approach to motivate the
use of LML layers for maximum recall.
Given access to the ground-truth data in addition
to the observation and assuming label independence,
we could maximize
the likelihood of a parametric model with
\begin{equation}
  P(Y, Y^\star\mid x) = \prod_{j\in \Y} P(j\in Y^\star\mid x).
  \label{eq:ml-likelihood}
\end{equation}
We can decompose $P(Y, Y^\star\mid x)$ as
\begin{equation*}
  \begin{split}
  P(Y, Y^\star\mid x) = &\prod_{j\in Y^\star} P(j\in Y^\star\mid x)
    \prod_{j\in \Y-Y^\star} P(j\not\in Y^\star\mid x). \\
    &\hspace{-13mm}\overbrace{
      \prod_{j\in Y} P(j\in Y^\star\mid x)
      \prod_{j\in Y^\star-Y} P(j\in Y^\star\mid x)}
  \end{split}
\end{equation*}
The difficulty in modeling this problem given
only the observed labels $Y$ comes from not knowing which
of the \emph{unobserved} labels should be
active or inactive.
In the case when all $|Y^\star|=k$, then the ground-truth
labels can be interpreted as vertices of the
LML polytope that have a value of 1 if the label
is present and 0 otherwise.
Thus, we can use a model that makes a prediction
on the LML polytope $f_\theta:\X\rightarrow\LML_k$.
The outputs of this model $\hat p = f_\theta(x)$
are then the likelihoods $\hat p_j \triangleq P(j\in Y^\star \mid x)$.
For example, $f_\theta$ can be modeled with a standard
deep feed-forward network with an LML layer at the end.
The set of predicted labels can be obtained with
$$\hat Y(x) = \{j \mid f_{\theta}(x)_j \geq
\pi\left(f_{\theta}(x)\right)_k\},$$
breaking ties if necessary in the unlikely case
that multiple
$f_{\theta}(x)_j = \pi\left(f_{\theta}(x)\right)_k$.
We next state assumptions under which we can
reason about maximum-likelihood solutions.

\begin{algorithm}[t]
  \caption{Maximizing top-$k$ recall via
  maximum likelihood with the LML layer.}
\label{alg:topk}
\textbf{Model:} $f_\theta: \X\rightarrow\RR^n $ \\
\textbf{Model Predictions:}
  $\hat Y_i = \{j \mid f_{\theta}(x_i)_j \geq
  \pi\left(f_{\theta}(x_i)\right)_k\}$ \\
\textbf{Training Procedure:}
\begin{algorithmic}[1]
  \While {unconverged}
    \State Sample $(x_i,Y_i)\sim \D$
    \State $\hat p = \Pi_{\LML_k}\left(f_\theta(x_i)\right)$
    \State Update $\theta$ with a gradient step
      $\nabla_\theta \ell(Y_i, \hat p)$ where
      \begin{equation*}
        \begin{split}
          \ell(Y_i, \hat p) = -\sum_{j\in Y_i} \log \hat p_j
        \end{split}
      \end{equation*}
  \EndWhile
\end{algorithmic}
\end{algorithm}

\textbf{Assumptions.}
For the following, we assume that
1) in the infinite data setting, the ground-truth labels
are able to be reconstructed from the observed labels
(e.g.~for a fixed feature, the observed labels are
sampled from the ground-truth labels with a non-zero
weight on each label),
2) there is no noise in the data generating process,
3) the true model is realizable and therefore
maximizing the likelihoods can be done exactly,
and
4) all $|Y^\star_i|=k$.
We claim that all of these assumptions can be reasonably
relaxed and we empirically show that LML layers are
effective in settings where these don't hold.

\begin{proposition}
  \label{prop:ml}
  Maximizing the likelihood of \mbox{$f_\theta(x_i):\X\rightarrow\LML_k$}
  on only the observed data
  $$\max_\theta\; \E\left[\prod_{j\in Y_i} (f_\theta(x_i))_j \right] \triangleq
  \E\left[ \prod_{j\in Y_i} P(j\in Y_i^\star \mid x_i) \right]$$
  implicitly maximizes
  $\E\left[ P(Y_i^\star \mid x_i) \right]$.
  All expectations are done over samples from the data
  generating process $(x_i, Y_i) \sim \D$.
\end{proposition}

This can be proven by observing that the model's LML
output space will allow the unobserved positive labels to
have high likelihood
$$\prod_{j\in Y^\star-Y} P(j\in Y^\star \mid x)$$
while forcing all the true negative data to have
low likelihood
$$\prod_{j\in \Y-Y^\star} P(j\in Y^\star \mid x).$$

We note that \cref{prop:ml} \emph{does not hold} for
a standard multi-label prediction model that
makes predictions onto the unit hypercube
$f_\theta: \X\rightarrow [0,1]^{n}$
where $$\hat p_j = f_\theta(x_i) \triangleq P(j\in Y^\star \mid x)$$
as only maximizing
$$\prod_{j\in Y_i} P(j\in Y^\star \mid x)$$ will result
in a collapsed model that predicts $\hat p_j=1$
for every label $j\in\Y$.

\begin{corollary}
  Maximizing the likelihood of $f_\theta:\X\rightarrow\LML_k$ on
  the observed data $\E\left[ P(Y_i\mid x_i) \right]$
  maximizes the recall of the observed data
  $\E\left[ {\rm recall}(Y, \hat Y)\right]$.
\end{corollary}

The ground-truth data are vertices of the LML polytope
and $f_\theta$ approaches the ground-truth likelihoods.
Thus the model's prediction $\hat Y(x)$ is the ground-truth
and the recall of the observed data is maximized.
We again note that the model's 0-1 error on the observed data
${\rm error}(Y, \hat Y)$ is in general \emph{not} minimized,
but that the error on the ground-truth data
${\rm error}(Y^\star, \hat Y)$ is minimized,
as the observed data may not have all of the labels that are
present in the ground-truth data.

We propose a gradient-based approach of solving this maximum
likelihood problem in \cref{alg:topk} that we use for
all of our experiments.

\subsection{Top-$k$ Image Classification}
\label{sec:lml:topk:image}

In top-$k$ image classification, the dataset consists
of images $x_i$ with single labels $y_i$ and the task
is to predict a set of $k$ labels $\hat Y$ that
maximizes ${\rm recall}(\{y_i\}, \hat Y)$.
We show in \cref{sec:lml:cifar} that LML models are competitive with
the state-of-the-art methods
for top-$k$ image classification on the noisy variant of
\mbox{CIFAR-100} from \citet{berrada2018smooth}.

\subsection{Scene Graph Generation}
\label{sec:lml:topk:sg}

As briefly introduced in \cref{sec:lml:rw:sg}, scene graph generation
methods take an image as input and output a graph of the objects
in the image (the nodes of the graph) and the relationships
between them (the edges of the graph).
One of the recent state-of-the-art methods that
is characteristic of many of the other methods is
Neural Motifs \citep{zellers2018neural}.
Neural Motifs and related models such as
\citet{xu2017scene}
make an assumption that
the relationships on separate edges are independent from
each other.
In this section, we show how we can use the maximum recall training
with an LML layer to make a minor modification
to the training procedure of these models that allows us
to relax this assumption with negligible computational overhead.

Specifically, the Neural Motifs architecture decomposes the scene
graph generation task as
\begin{equation*}
P(G \mid I) = P(B \mid I) \ P(O \mid B, I) P(R \mid B, O, I)
\end{equation*}
where $G$ is the scene graph, $I$ is the input image,
$B$ is a set of region proposals,
and $O$ is a set of object proposals.
The relationship generation process $P(R\mid B, O, I)$
makes an independence assumption that,
given a latent variable $z$ that is present at
each edge as $z_{ij}$,
the relationships on each edge are independent.
That is,
$$P([x_{i\to j}]_{ij}\mid z, B,O, I)=\prod_{i,j} P(x_{i\to j}\mid z_{ij},B,O,I),$$
where the set of relationships between
all of the nodes is $R=[x_{i\to j}]_{ij}$.

Neural Motifs models these probabilities with
\begin{equation}
  \label{eq:sg:vanilla}
  P(x_{i \to j} \mid B, O, I) =
  \hat p_{ij} \triangleq {\rm softmax}(z_{ij}) \in\Delta_n,
\end{equation}
where $n$ is the number of relationships for the task.
The predictions are made in the $n$-simplex instead
of the $(n-1)$-simplex because an additional class
is added to indicate that no relationship is present
on the edge.
For inference, graphs are generated by selecting
the relationships that have the highest probability by
concatenating all $p_{ij}$ and selecting the top $k$.
Typical values of $k$ are 20, 50, and 100.
The method is then evaluated on the top-$k$ recall
of the scene graphs; i.e.~the number of ground-truth
relationships that are in the model's top-$k$
relationship predictions.

Two drawbacks of the vanilla Neural Motif model of treating
the edge relationships as independent softmax functions
are that
1) edges with multiple relationships will never
achieve perfect likelihood because the softmax function
is being used to make a prediction at each edge.
If multiple relationships are present on a single edge,
the training code for Neural Motifs randomly
samples a single one to use for the update in that iteration.
For inference, multiple relationships on a node
\emph{can} be predicted if their individual probabilities
are within the top-$k$ threshold, although they are
still subject to the simplex constraints and therefore
may be unreasonably low; and
2) the evaluation metric of generating a graph
with $k$ relationships is not part of the training
procedure that just treats each edge as a classification
problem that maximizes the likelihood of the observed
relationships.

Using an LML layer to predict all of the relationship
probabilities jointly overcomes these drawbacks.
We model the joint probability with
\begin{equation}
  \label{eq:sg:lml}
  P([x_{i \to j}]_{ij} \mid z, B, O, I) = \Pi_{\LML_k}\left({\rm cat}([z_{ij}]_{ij})\right)
\end{equation}
where ${\rm cat}$ is the concatenation function.
This is now a top-$k$ recall problem that we train by
maximizing the likelihood of the observed relationships
with \cref{alg:topk}.
We have added the LML training procedure to the official
Neural Motifs codebase in $\approx$20 lines of code to
project $[z_{ij}]_{ij}$ onto the LML polytope instead
of projecting each $z_{ij}$ onto the simplex, and
to optimize the likelihood of the data jointly
instead of independently.

The LML approach for scene graph generation overcomes
both of the drawbacks of the vanilla approach by
1) allowing the ground-truth data to achieve
near-perfect likelihood as multiple relationships
are allowed to be present between the edges, and
2) introducing the knowledge predicting
$k$ nodes into the training procedure.
One downside of the LML approach for scene graph
generation is that the training procedure now depends
on $k$ while the vanilla training procedure does not.
We empirically show that it is typically competitive to
train with a fixed $k$ and evaluate for others.

\begin{figure*}[!t]
  \centering
  \includegraphics[width=.45\textwidth]{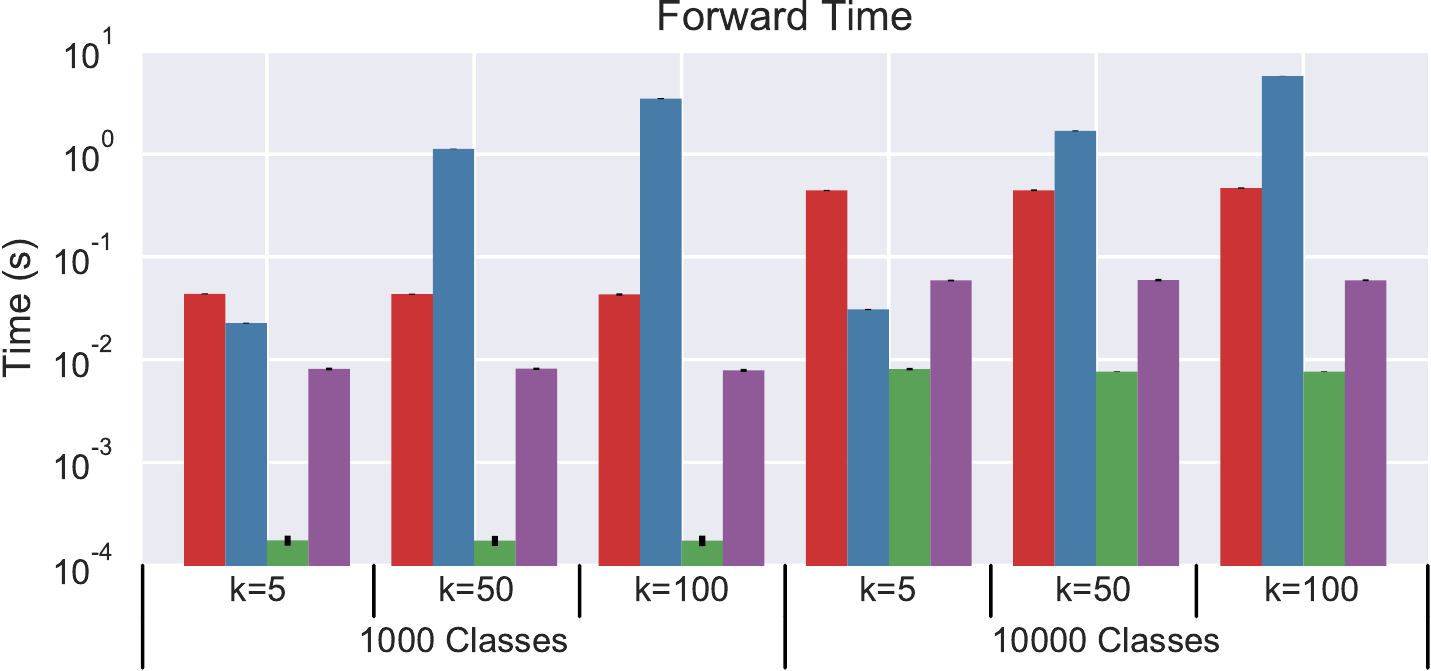}
  \includegraphics[width=.45\textwidth]{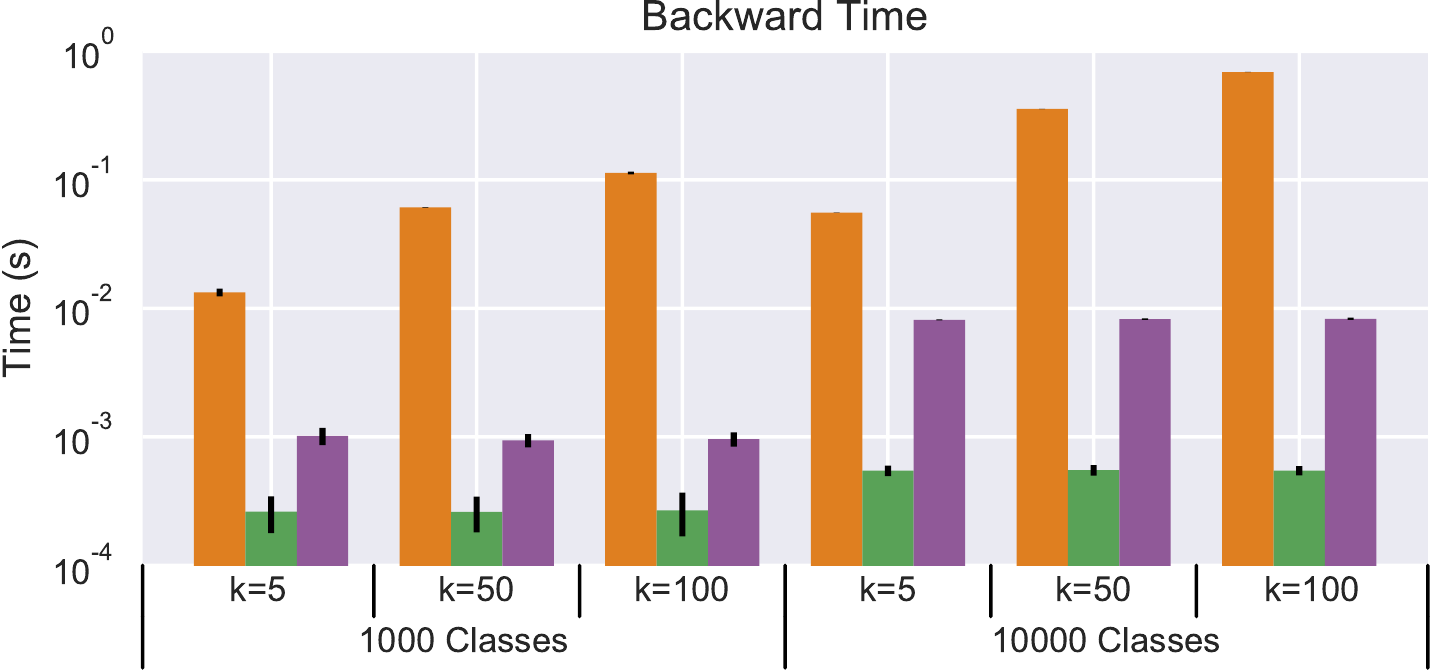}

  Smooth SVM (%
    \cblock{187}{64}{60} SA \enskip
    \cblock{83}{123}{164} DC \enskip
    \cblock{210}{132}{57} Backward%
  ) \qquad
  \cblock{106}{160}{95} Ent$_{\rm tr}$ \qquad
  \cblock{159}{92}{149} LML
  \caption{
    Timing performance results. Each point is from 50 trials
    on an unloaded system.
  }
  \label{fig:perf}
\end{figure*}

\newpage
\section{Experimental Results}
\label{sec:lml:ex}

In this section we study the computational efficiency of
the LML layer and show that it performs competitively with
other methods for top-$k$ image classification.
When added to the Neural Motifs model for scene graph
generation, LML layers improve the modeling capability
with almost no computational overhead.

We have released a PyTorch implementation of
the LML layer and our experimental code at: \\[2mm]
\centerline{\url{https://github.com/locuslab/lml}}

\subsection{Performance Comparisons}
\label{sec:lml:perf}

The LML layer presented in \cref{mod:lml} has a non-trivial
forward and backward pass that may be computationally
expensive if not implemented efficiently.
To better understand the computational costs of the LML layer,
we have measured the timing performance of our layer
in comparison to the Smooth SVM loss from \citet{berrada2018smooth}
and the truncated top-$k$ entropy Ent$_{\rm tr}$
from \citet{lapin2016loss},
which we review in \cref{sec:lml:entr-derivation}.
The Summation Algorithm (SA) and Divide-and-Conquer (DC)
algorithms for the Smooth SVM loss are further described in
\citet{berrada2018smooth}.
We use the official Smooth SVM implementation and
have re-implemented the truncated top-$k$ entropy
in PyTorch for our experiments.
The truncated top-$k$ entropy loss function is only
bottlenecked by a sorting operation, which we
implemented using PyTorch's sort function.

\cref{fig:perf}
measures the performance of our method in comparison to the
Smooth SVM and truncated top-$k$ entropy using
the profiling setup from
\citet{berrada2018smooth}.
We use a minibatch size of 256
and runs 50 trials for each data point.
We ran all of the experiments on an unloaded
NVIDIA GeForce GTX 1080 Ti GPU.
The forward pass of the smooth SVM becomes computationally
expensive as $k$ grows while the LML layer's performance
and the truncated top-$k$ entropy method's performance
remain constant.
The top-$k$ entropy loss is only bottlenecked by a
sorting operation and significantly outperforms
both the Smooth SVM and LML layers.
We emphasize that \citet{berrada2018smooth}
did not consider the truncated top-$k$ entropy
method as a baseline.

\subsection{Top-$k$ Image Classification on CIFAR-100}
\label{sec:lml:cifar}

\begin{figure*}[!t]
  \centering
  \includegraphics[width=.45\textwidth]{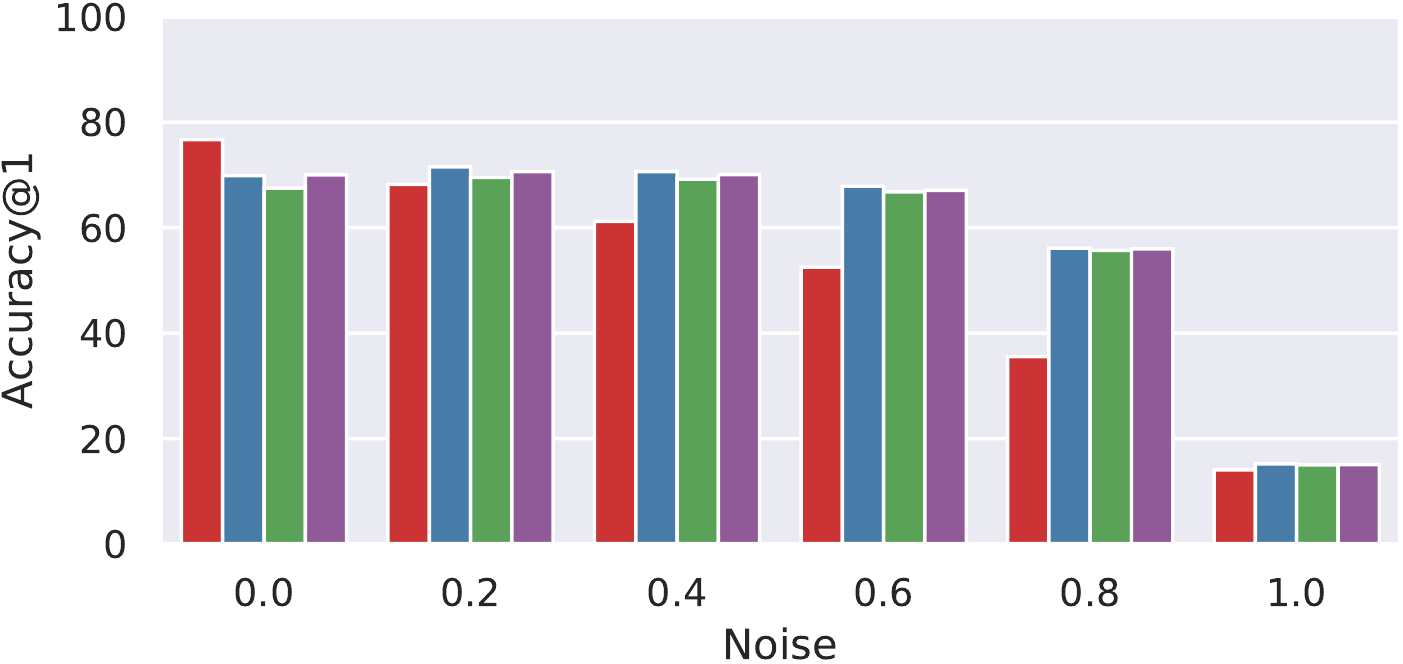}
  \includegraphics[width=.45\textwidth]{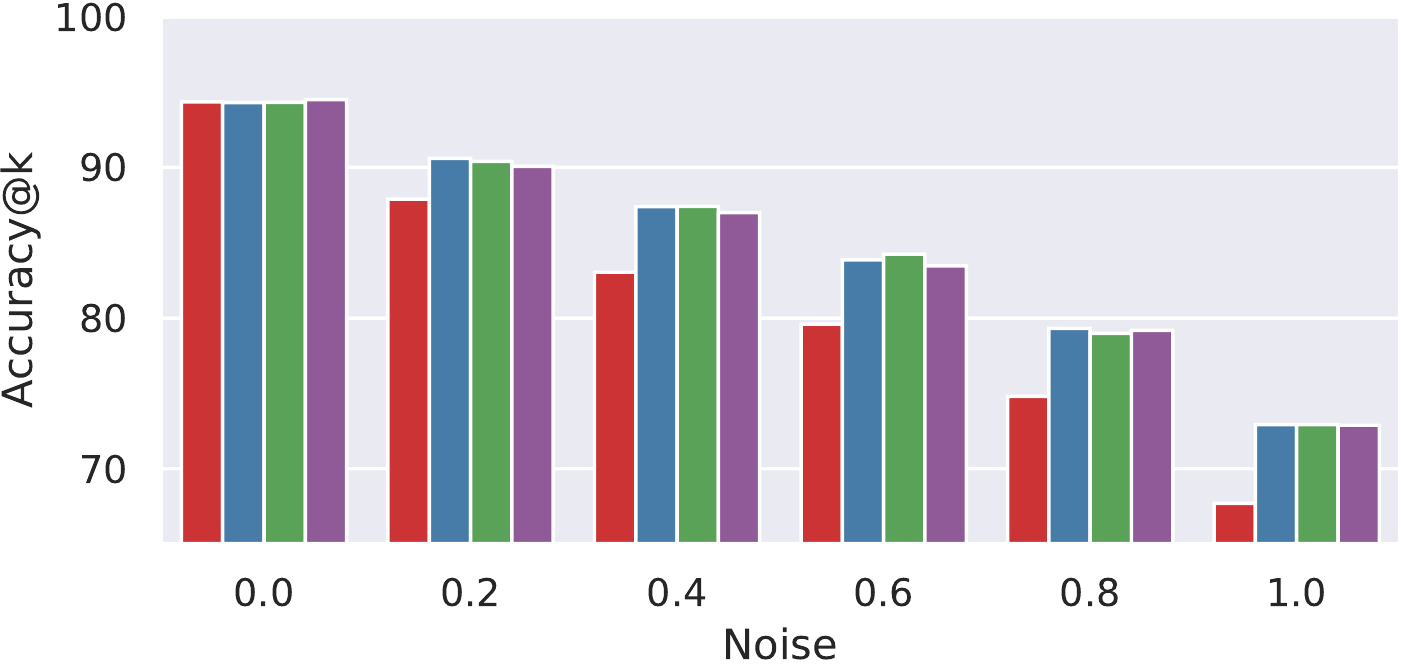}

  \cblock{187}{64}{60} Cross-Entropy \enskip
  \cblock{83}{123}{164} Smooth SVM \enskip
  \cblock{106}{160}{95} Ent$_{\rm tr}$ \enskip
  \cblock{159}{92}{149} LML
  \caption{
    Testing performance on CIFAR-100 with label noise.
  }
  \label{fig:cifar-topk}
\end{figure*}

We next evaluate the LML layer on the noisy top-5 CIFAR-100 task
from \citet{berrada2018smooth} that uses
the DenseNet 40-40 architecture \citep{huang2017densely}.
The CIFAR-100 labels are
organized into 20 ``coarse''
classes, each consisting of 5 ``fine'' labels.
With probability $p$, noise is added to the labels
by resampling from the set of ``fine'' labels.

\cref{fig:cifar-topk} shows that the LML model is competitive
with the other baseline methods for this task:
standard cross-entropy training, the Smooth SVM models,
and the truncated entropy loss.
We used the experimental setup and code from
\citet{berrada2018smooth} and added the LML
experiments with a few lines of code.
Notably, we also re-implemented the truncated entropy
loss from \citet{lapin2016loss} as another reasonable
baseline for this task, which \citet{berrada2018smooth}
did not consider as a baseline.
Following the method of \citet{berrada2018smooth},
we ran four seeds for the truncated entropy and
LML models and report the average test performance.
For reference, a model making random predictions
would obtain 1\% top-1 accuracy and 5\% top-5 accuracy.

The results show that relative to the cross-entropy,
the smooth SVM, truncated entropy, and LML losses
perform similarly.
Relative to each other the best method is not clear,
which is consistent with the experimental results
on other tasks in \citet{lapin2016loss}.
We interpret these results as showing that all of
the methods evaluated for top-$k$ optimization
learn nearly identical models despite being
formulated differently.

\begin{figure*}[!t]
  \centering
  \includegraphics[width=.45\textwidth]{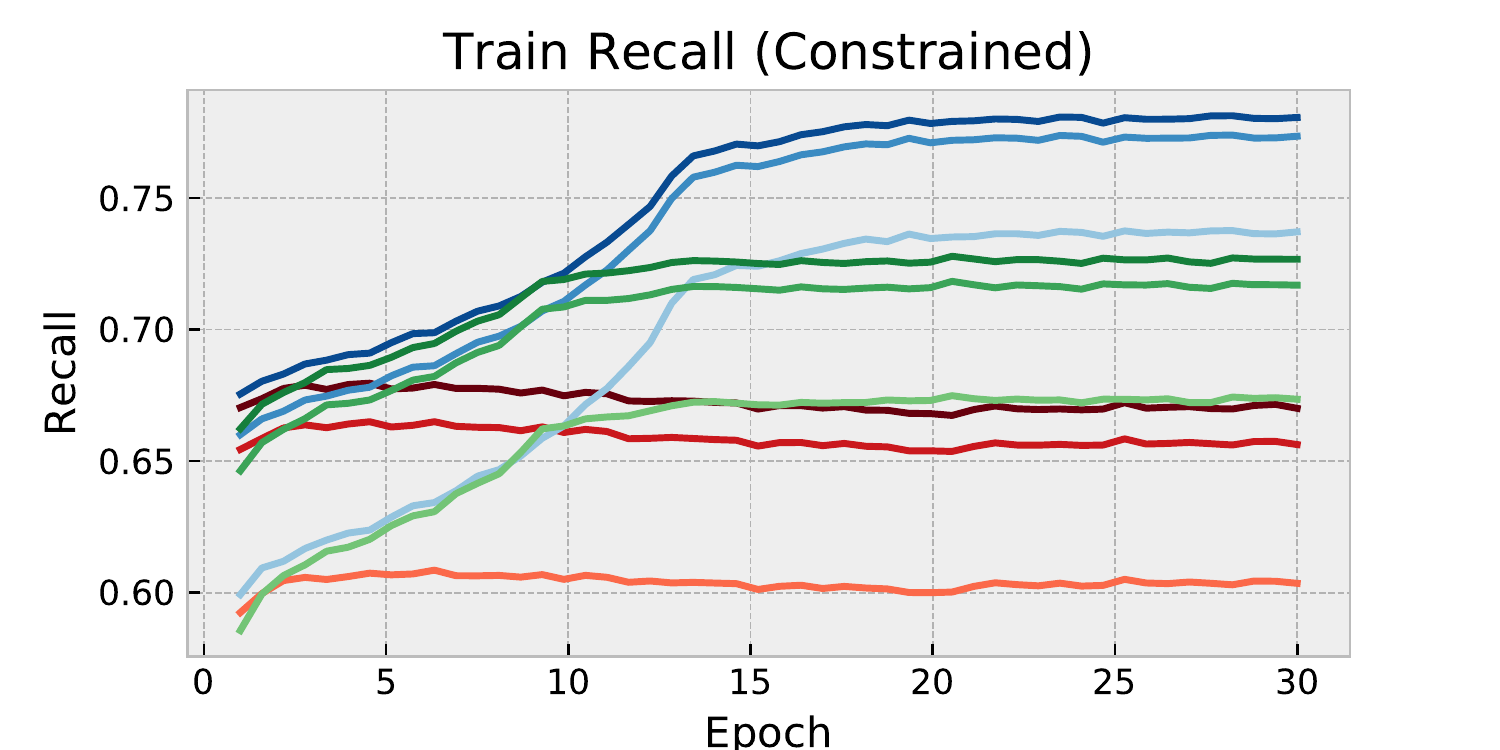}
  \includegraphics[width=.45\textwidth]{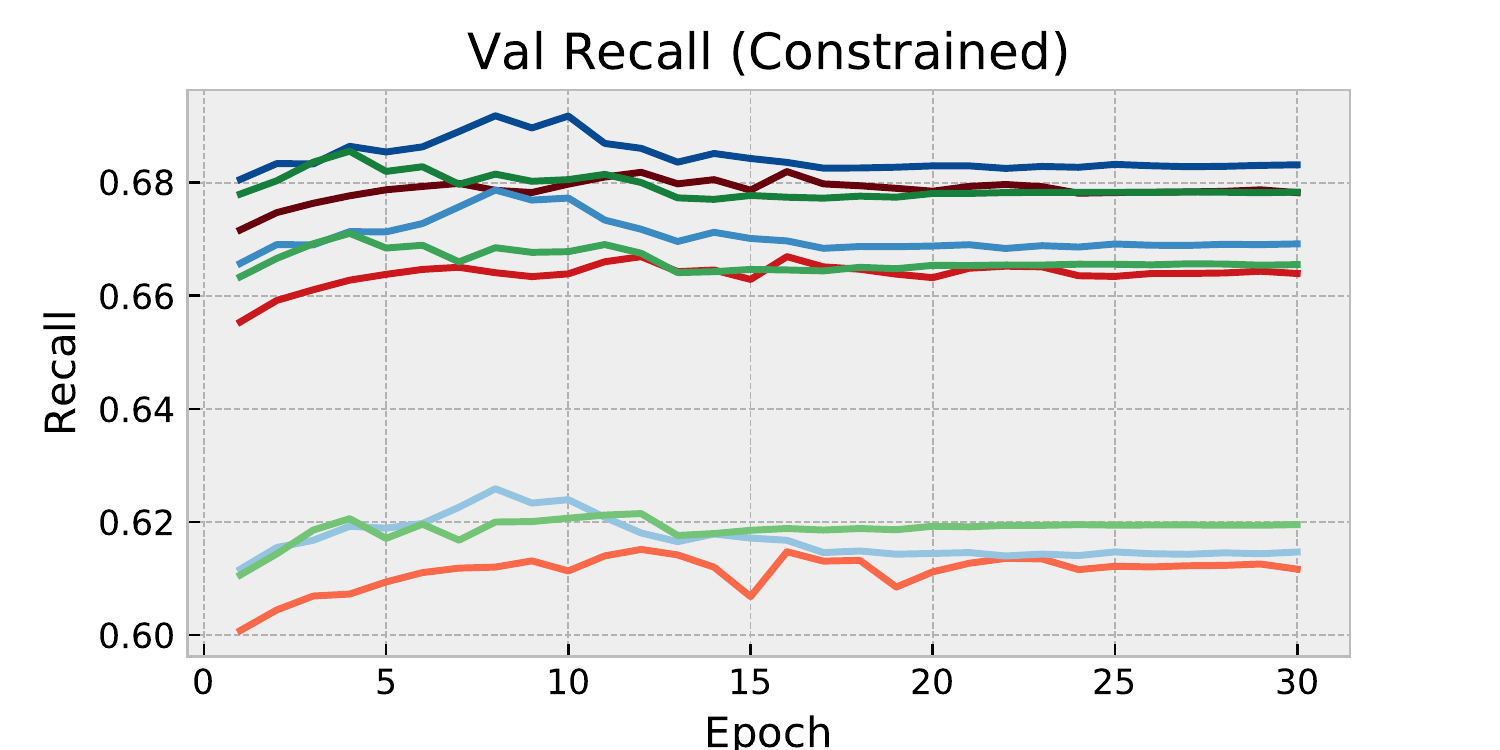}

  \footnotesize
  \citep{zellers2018neural} R@(%
    \cblock{234}{114}{84} 20 \hspace{0.1em}
    \cblock{186}{47}{41} 50 \hspace{0.1em}
    \cblock{94}{15}{18} 100%
  )\hspace{5mm}
  +Ent$_{\rm tr}$ R@(%
    \cblock{134}{194}{126} 20 \hspace{0.1em}
    \cblock{90}{162}{96} 50 \hspace{0.1em}
    \cblock{59}{125}{67} 100%
  )\hspace{5mm}
  +LML R@(%
    \cblock{158}{195}{220} 20 \hspace{0.1em}
    \cblock{81}{137}{190} 50 \hspace{0.1em}
    \cblock{31}{73}{141} 100%
  )
  \caption{
    (Constrained) scene graph generation on the Visual Genome:
    Training and validation progress comparing the vanilla
    Neural Motif model to the Ent$_{\rm tr}$ and LML versions.
  }
  \label{fig:sg-training-cons}
\end{figure*}

\begin{figure*}[!t]
  \centering
  \includegraphics[width=.45\textwidth]{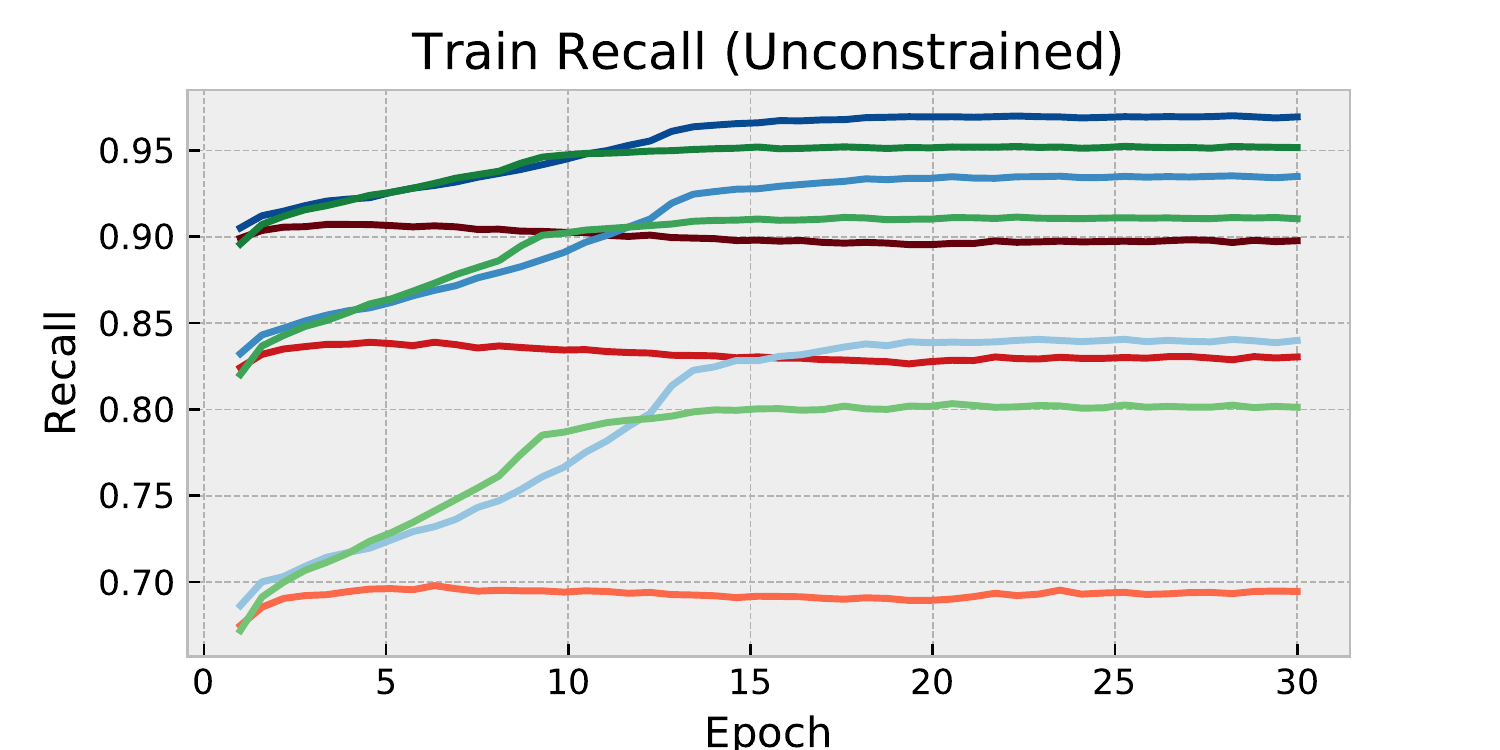}
  \includegraphics[width=.45\textwidth]{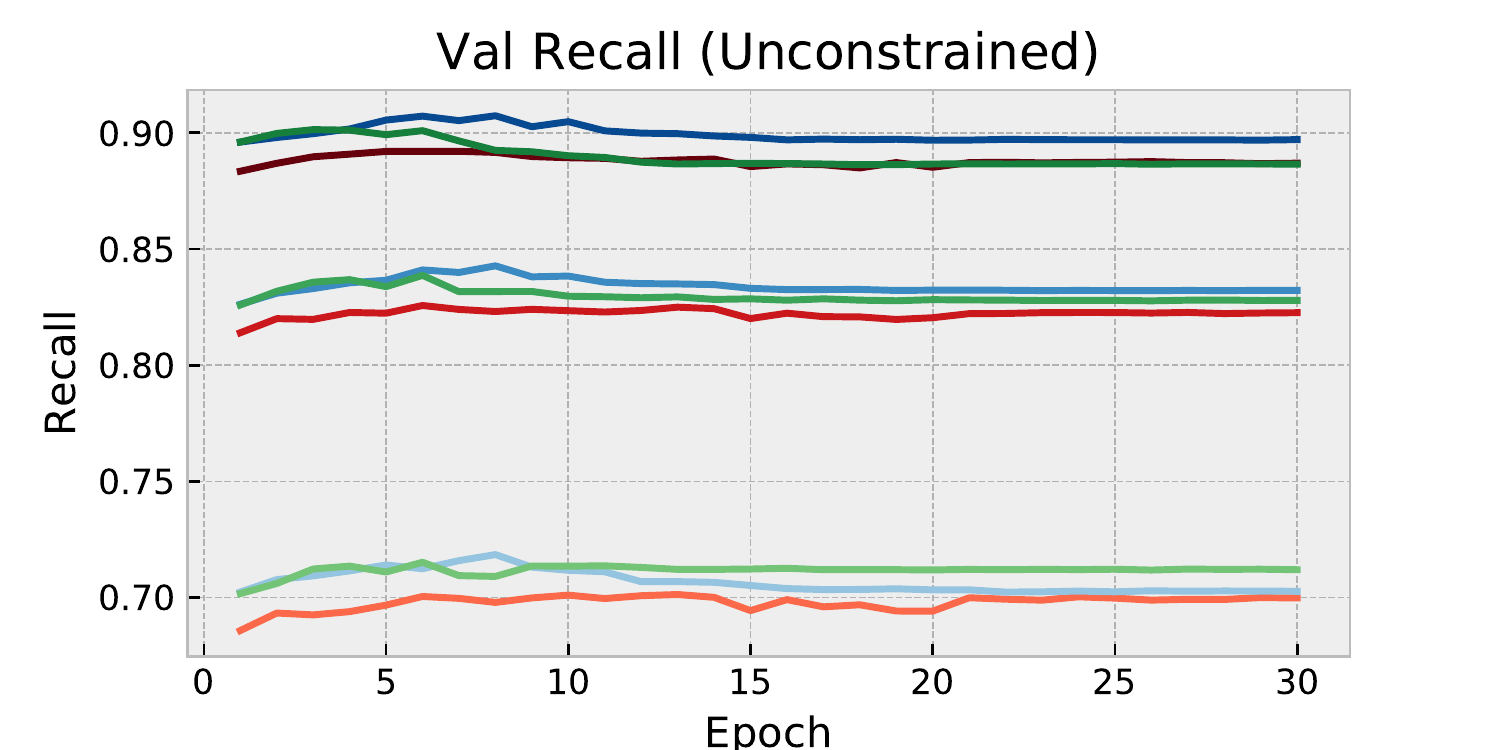}

  \footnotesize
  \citep{zellers2018neural} R@(%
    \cblock{234}{114}{84} 20 \hspace{0.1em}
    \cblock{186}{47}{41} 50 \hspace{0.1em}
    \cblock{94}{15}{18} 100%
  )\hspace{5mm}
  +Ent$_{\rm tr}$ R@(%
    \cblock{134}{194}{126} 20 \hspace{0.1em}
    \cblock{90}{162}{96} 50 \hspace{0.1em}
    \cblock{59}{125}{67} 100%
  )\hspace{5mm}
  +LML R@(%
    \cblock{158}{195}{220} 20 \hspace{0.1em}
    \cblock{81}{137}{190} 50 \hspace{0.1em}
    \cblock{31}{73}{141} 100%
  )
  \caption{
    (Unconstrained) scene graph generation on the Visual Genome:
    Training and validation progress comparing the vanilla
    Neural Motif model to the Ent$_{\rm tr}$ and LML versions.
  }
  \label{fig:sg-training-uncon}
\end{figure*}

\subsection{Scene Graph Generation}

\begin{table*}[t]
\footnotesize
\centering
\begin{tabular}{@{}c@{\hspace{0.4em}} l c@{\hspace{0.2em}} c@{\hspace{4em}}cc c@{\hspace{0.2em}} c@{\hspace{4em}}cc c@{}}
\toprule
      && \phantom{} & \multicolumn{3}{c}{Predicate Classification (Constrained)} &  \phantom{} & \multicolumn{3}{c}{Predicate Classification (Unconstrained)} \\
    \cmidrule{4-6} \cmidrule{8-10}
& Model && R$@$20 & R$@$50  & R$@$100 && R$@$20 & R$@$50  & R$@$100 \\
\midrule
      & \citep{zellers2018neural} && 61.5 & 66.7 & 68.2 &&
                                70.1 & 82.6 & 89.2 \\ \midrule
      & +LML-20 && {\bf 62.6} & {\bf 67.9} & {\bf 69.2} &&
                           {\bf 71.9} & {\bf 84.3} & {\bf 90.7} \\
      & +LML-50 && {\bf 62.5} & {\bf 67.8} & {\bf 69.1} &&
                           71.6 & 84.1 & 90.5 \\
      & +LML-100 && 61.2 & 66.3 & 67.7 &&
                            70.4 & 83.3 & {\bf 90.7} \\ \midrule
      & +Ent$_{\rm tr}$-20 && 62.1 & 67.1 & 68.6 &&
                                     71.5 & 83.9 & 90.1 \\
      & +Ent$_{\rm tr}$-50 && 61.7 & 66.9 & 68.4 &&
                                     71.1 & 84.0 & 90.3 \\
      & +Ent$_{\rm tr}$-100 && 60.7 & 66.3 & 67.8 &&
                                      69.7 & 83.5 & 90.1 \\
\bottomrule
\end{tabular}
\caption{
  Scene graph generation on the Visual Genome: Best Validation Recall Scores
}
\label{tab:sg-results-val}
\end{table*}

\begin{table*}[t]
\centering
\footnotesize
\begin{tabular}{@{}c@{\hspace{0.4em}} l c@{\hspace{0.2em}} c@{\hspace{4em}}cc c@{\hspace{0.2em}} c@{\hspace{4em}}cc c@{}}
\toprule
      && \phantom{} & \multicolumn{3}{c}{Predicate Classification (Constrained)} &  \phantom{} & \multicolumn{3}{c}{Predicate Classification (Unconstrained)} \\
    \cmidrule{4-6} \cmidrule{8-10}
& Model && R$@$20 & R$@$50  & R$@$100 && R$@$20 & R$@$50  & R$@$100 \\
\midrule
      & \citep{zellers2018neural} && 58.5 & 65.2 & 67.1
  && {\bf 66.6} & {\bf 81.1} & {\bf 88.2} \\ \midrule
      & +Ent$_{\rm tr}$ && {\bf 59.4} & {\bf 66.1} & {\bf 67.8}
  && 60.8 & 70.7 & 75.6 \\
      & +LML && 58.5 & {\bf 66.0} & {\bf 67.9}
  && 64.2 & 79.4 & 87.6 \\
\bottomrule
\end{tabular}
\caption{
  Scene graph generation on the Visual Genome: Test Dataset Results.
}
\label{tab:sg-results}
\end{table*}

For our scene graph generation experiments we use the
\textsc{MotifNet-LeftRight} model, experimental setup,
and official code from \citet{zellers2018neural}.
We added the LML variant with $\approx$20 lines of code.
This experiment uses the Visual Genome dataset
\citep{krishna2017visual}, using the
the publicly released preprocessed data and
splits from \citet{xu2017scene}.
In this report, we focus solely on the
\emph{Predicate Classification} evaluation mode
\verb!PredCls!
which uses a pre-trained detector and classifier and
only measures improvements to the relationship
predicate model $P(R\mid B, O, I)$.
Our methods can also be extended to the other evaluation
modes that jointly learn models for the detection and
object classification portions $P(G \mid I)$ and
we believe that our improvements on the
\verb!PredCls! mode upper-bound the improvements
an LML layer would add to the other evaluation modes.
\emph{Constrained} graph generation constrains the
graphs to have at most a single relationship present
at each edge, and is more common in the literature.

We also consider using a modified version of
the truncated top-$k$ entropy loss that
we derive in \cref{sec:lml:entr-ml-derivation}.
We do not consider modifications of the Smooth SVM
because the performance results in \cref{sec:lml:perf}
show that the approach is
nearly computationally infeasible when
scaling to the size necessary for scene-graph generation.
An image with 20 objects and 50 possible relationships
generates $20(19)(50)=19000$ possible relationship candidates.

All of the LML and truncated top-$k$ entropy (Ent$_{\rm tr}$) models
we evaluate in this section are trained on predicting
graphs with 20
relationships, which perform competitively on the
validation dataset.
\cref{fig:sg-training-uncon} shows the training progress for
unconstrained graph generation.
\cref{tab:sg-results-val} shows the validation performance
for the truncated top-$k$ entropy and LML layers when trained
for $k\in\{20,50,100\}$.
\cref{fig:sg-training-cons} shows that the truncated
top-$k$ entropy and LML approach both add representational
capacity and improve the training recall by 5-10\%
for all evaluation modes for constrained graph generation.
This behavior is also present for unconstrained graph
generation in \cref{fig:sg-training-uncon}.
These improvements are not as significant on
the validation dataset, or on the test dataset
in \cref{tab:sg-results}.
In the unconstrained evaluation mode, the LML layers
outperform the truncated top-$k$ entropy and almost
reach the performance of the baseline.
This performance gap is likely because the Visual Genome
dataset has a lot of noise from the human-generated scene graph
annotations, and the LML model fits to more noise
in the training dataset that does not generalize
to the noise present in the validation or test datasets.
Surprisingly, the LML model improves the constrained
graph generation test performance but slightly decreases
the unconstrained graph generation performance.
We theorize this is because of noise that the
model starts to overfit to and that constraining
the model to only make a single prediction at each
edge is a reasonable heuristic.

\section{Conclusions}
We have presented the LML layer for top-$k$ multi-label learning.
The LML layer has a forward pass that can be efficiently
computed with a parallel bracketing method and a backward
pass that can be efficiently computed by perturbing
the KKT conditions of the optimization problem.
We have empirically demonstrated
that the LML layer adds representational capacity
for top-$k$ optimization and in many cases can be
added to existing code with $\approx$20 additional
lines of code.
As a compelling future research direction for these layers,
these layers can also enable deep structured
prediction models to be used for top-$k$ prediction.

\subsection*{Acknowledgments}
We thank Rowan Zellers for help reproducing and running
the Neural Motifs training code and Mathieu Blondel
and Andr{\'e} Martins for useful comments.

\newpage
\printbibliography

\newpage
\appendix
\section{Truncated Top-$k$ Entropy Derivation}
\label{sec:lml:entr-derivation}

This section reviews the truncated top-$k$ entropy derivation
from Section~2.5 of \citet{lapin2016loss}.
We start with the standard likelihood
\begin{equation}
  P(y\mid x)=\frac{\exp\{f_y(x)\}}{\sum_j \exp\{f_j(x)\}}
\end{equation}
and then consider the negative log-likelihood
\begin{equation}
  \begin{split}
    -\log P(y\mid x) &= -\log \frac{\exp\{f_y(x)\}}{\sum_j \exp\{f_j(x)\}} \\
    &= \log\frac{\sum_j \exp\{f_j(x)\}}{\exp\{f_y(x)\}} \\
    &= \log \left( 1 + \sum_{j\neq y} \exp\{f_j(x)-f_y(x)\} \right)
  \end{split}
\end{equation}
Truncating the index set of the last sum gives the truncated
top-$k$ entropy loss
\begin{equation}
  \log \left( 1 + \sum_{j\in \JJ_y} \exp\{f_j(x)-f_y(x)\} \right)
\end{equation}
where $\JJ_y$ are the indices of the $m-k$ smallest
components of $\left(f_j(x)\right)_{j\neq y}$.
This loss is small whenever the top-$k$ error is zero.

\newpage
\section{Multi-Label Truncated Top-$k$ Entropy Derivation}
\label{sec:lml:entr-ml-derivation}

The truncated top-$k$ entropy loss from \citet{lapin2016loss}
is a competitive and simple loss function for optimizing
the model's top-$k$ predictions in single-label classification tasks.
In this section, we show how it can be extended to optimizing
the top-$k$ predictions in multi-label classification tasks,
such as scene graph generation.

We start by making an independence assumption between the
observed labels and decomposing the likelihood as
\begin{equation}
  P(Y\mid x)=\prod_i P(Y_i | x).
\end{equation}
Then, we can assume the likelihood of each label is
obtained with a softmax as
\begin{equation}
  \label{eq:entr-ml-label}
  P(Y_i\mid x)=\frac{\exp\{f_{y_i}(x)\}}{\sum_j \exp\{f_j(x)\}}.
\end{equation}
We note that in general, maximum-likelihood estimation
of the form \cref{eq:entr-ml-label} will never achieve
perfect likelihood as the softmax restricts the
likelihoods over all of the labels.
However following the approach from \citet{lapin2016loss},
we can rearrange the terms of the negative log-likelihood
and truncate parts of to obtain a reasonable loss function.

\begin{equation}
  \begin{split}
    -\log P(Y\mid x) &= -\sum_i \log \frac{\exp\{f_{y_i}(x)\}}{\sum_j \exp\{f_j(x)\}} \\
    &= \sum_i \log\frac{\sum_j \exp\{f_j(x)\}}{\exp\{f_{y_i}(x)\}} \\
    &= \sum_i \log \left( 1 + \sum_{j\neq y_i} \exp\{f_j(x)-f_{y_i}(x)\} \right)
  \end{split}
\end{equation}
Truncating the index set of the last sum gives the multi-label
truncated top-$k$ entropy loss
\begin{equation}
  \sum_i \log \left( 1 + \sum_{j\in \JJ_y} \exp\{f_j(x)-f_{y_i}(x)\} \right)
\end{equation}
where $\JJ_y$ are the indices of the $m-k$ smallest
components of $\left(f_j(x)\right)_{j\not\in y}$.
This loss is small whenever the top-$k$ recall
is zero.

\newpage
\section{The entropy surface of projections}
\label{app:entropies}

In this section we visualize the entropy penalties that the
LML and csoftmax projections use, which is inspired by
the visualizations in \citep[Appendix A.2]{blondel2019learning}.
To provide more intuition,
we also show the entropy penalties that the sigmoid
and softmax functions use.

\begin{figure}[t]
  \centering
  \includegraphics[height=34mm]{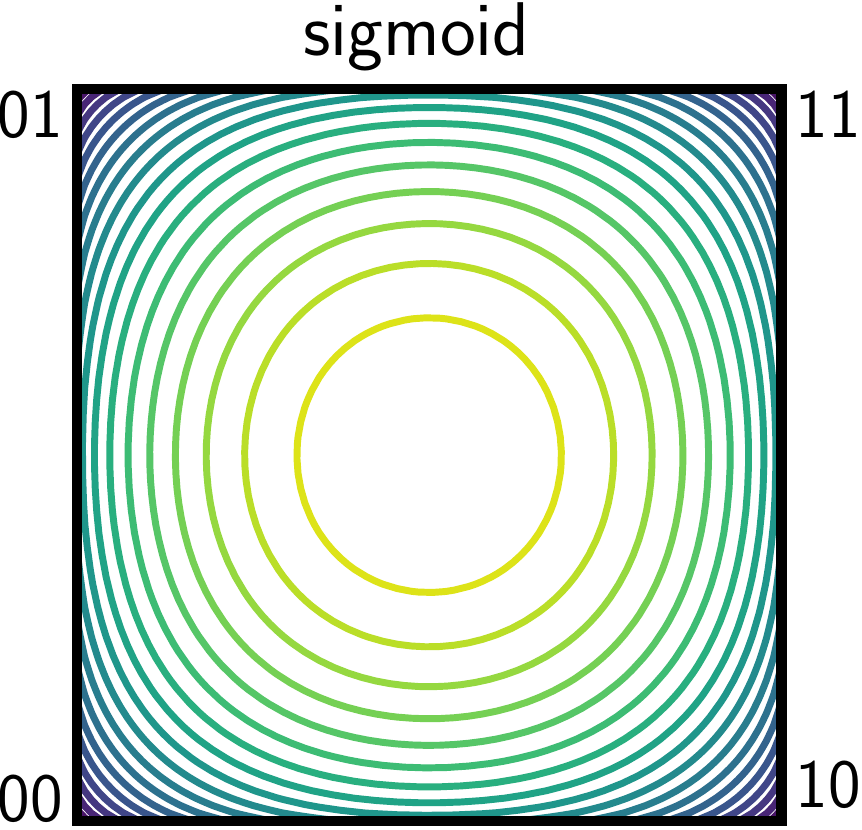}
  \hfill
  \includegraphics[height=34mm]{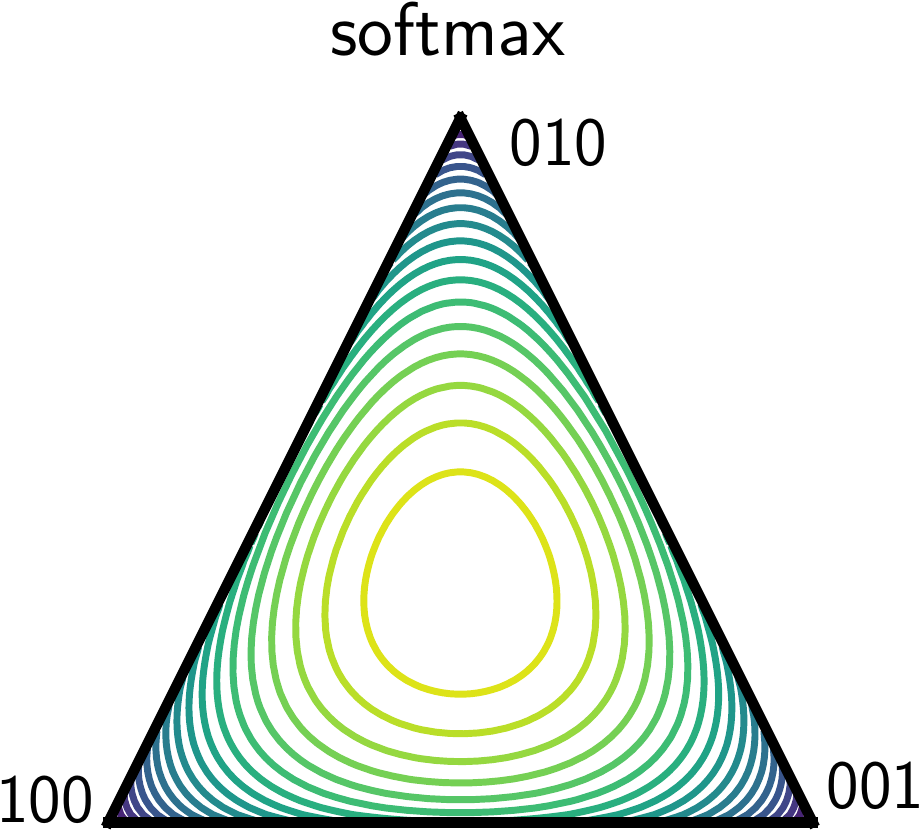} \\[3mm]
  \includegraphics[height=34mm]{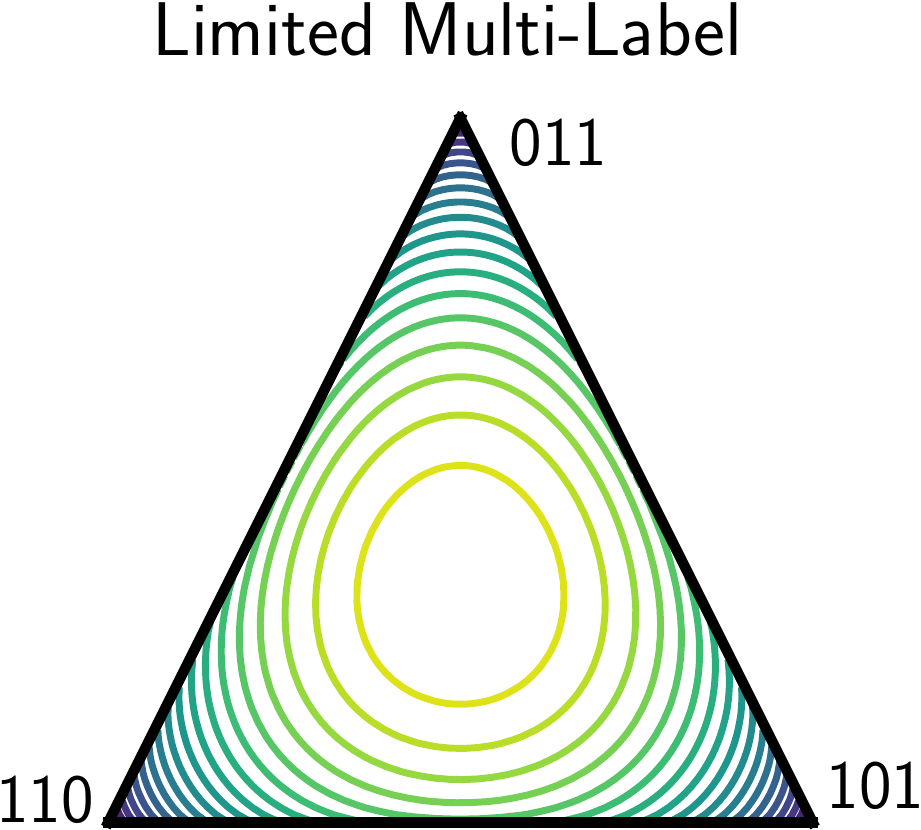}
  \hfill
  \includegraphics[height=34mm]{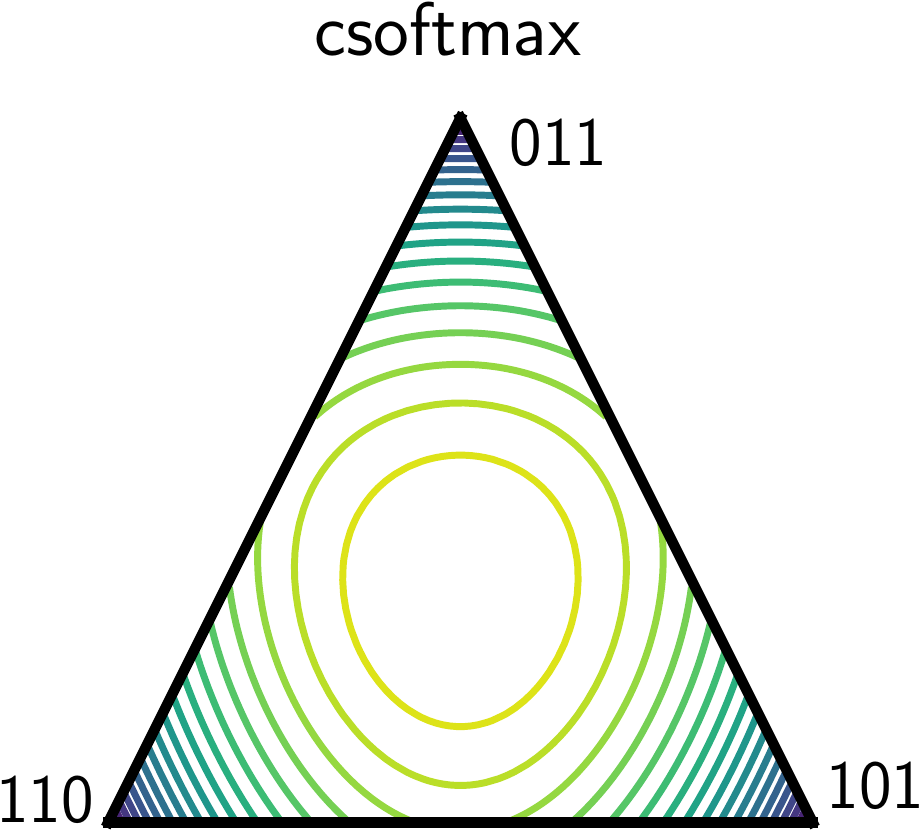}
  \caption{Comparison of entropy penalties of projections.
    The centers of these polytopes have the highest entropy and
    the vertices have the lowest.
    This illustrates our reasons for choosing the binary
    entropy penalty in the LML layer -- it provides a penalty
    surface that more closely resembles the sigmoid
    and softmax functions.
  }
  \label{fig:entr-comparisons}
\end{figure}

\cref{fig:entr-comparisons} shows the entropy surfaces over the
polytopes of the sigmoid and softmax, as well as the binary entropy
penalty from the LML projection in \cref{eq:lml-proj}, and the
unidirectional entropy penalty of the csoftmax when it is used
to project onto the LML polytope.

The following theorems review the optimization viewpoint of
the sigmoid function for multi-label classification and the
softmax function for single-label multi-class classification,
and are proved, \eg,
in \cite[Section 2.4]{amos2019differentiable}.

\begin{theorem}
  The sigmoid or logistic function, defined by $f(x) = (1+e^{-x})^{-1}$,
  can be interpreted as projecting a point $x\in\RR^n$ onto
  the interior of the unit hypercube as
  \begin{equation}
    f(x) = \argmin_{0<y<1} \;\; -x^\top y -H_b(y),
    \label{eq:sigmoid-proj}
  \end{equation}
  where $H_b(y) = - \left(\sum_i y_i\log y_i + (1-y_i)\log (1-y_i)\right)$ is the
  binary entropy function.
\end{theorem}

\begin{theorem}
  The softmax, defined by $f(x)_j = e^{x_j} / \sum_i e^{x_i}$,
  can be interpreted as projecting a point $x\in\RR^n$ onto
  the interior of the $(n-1)$-simplex
  $$\Delta_{n-1}=\{p\in\RR^n\; \vert\; 1^\top p = 1 \; \; {\rm and} \;\; p \geq 0 \}$$
  as
  \begin{equation}
    f(x) = \argmin_{0<y<1} \;\; -x^\top y - H(y) \;\; \st\;\; 1^\top y = 1
    \label{eq:simplex-proj}
  \end{equation}
  where $H(y) = -\sum_i y_i \log y_i$ is the entropy function.
\end{theorem}

\newpage\vfill~

\end{document}